\PassOptionsToPackage{table,xcdraw}{xcolor}
\documentclass[runningheads]{llncs}

\usepackage{eccv}

\usepackage{eccvabbrv}
\usepackage{graphicx}
\usepackage{booktabs}
\usepackage{multirow}
\usepackage{amsmath}
\usepackage{amssymb}
\usepackage{soul}
\usepackage{changepage,threeparttable}
\usepackage{mmstyle}
\usepackage{listings}
\usepackage{wrapfig}
\usepackage{lipsum}
\usepackage{titletoc}
\usepackage{longtable}
\usepackage{xcolor}
\usepackage{pdfpages}

\newcommand{\tablestyle}[2]{\setlength{\tabcolsep}{#1}\renewcommand{\arraystretch}{#2}\centering\small}
\definecolor{citecolor}{HTML}{0071BC}
\definecolor{bluecolor}{HTML}{0071BC}
\definecolor{redcolor}{HTML}{BD1C00}
\definecolor{greencolor}{HTML}{268E39}

\usepackage[accsupp]{axessibility}  

\usepackage{hyperref}

\usepackage{orcidlink}

\begin{document}

\title{PointLLM: Empowering Large Language Models to Understand Point Clouds} 

\author{Runsen Xu\inst{1,2} \and
Xiaolong Wang \inst{3} \and Tai Wang \inst{2\dag} \and Yilun Chen \inst{2} \and
Jiangmiao Pang\inst{2\dag} \and Dahua Lin \inst{1,2,4}}

\authorrunning{R.~Xu et al.}

\institute{The Chinese University of Hong Kong \and
Shanghai AI Laboratory \and Zhejiang University \and Centre for Perceptual and Interactive Intelligence
}

\renewcommand{\thefootnote}{\dag}
\footnotetext[1]{Corresponding authors: \{wangtai, pangjiangmiao\}@pjlab.org.cn}

\maketitle

\begin{abstract}
The unprecedented advancements in Large Language Models (LLMs) have shown a profound impact on natural language processing but are yet to fully embrace the realm of 3D understanding. This paper introduces PointLLM, a preliminary effort to fill this gap, empowering LLMs to understand point clouds and offering a new avenue beyond 2D data. PointLLM understands colored object point clouds with human instructions and generates contextually appropriate responses, illustrating its grasp of point clouds and common sense. Specifically, it leverages a point cloud encoder with a powerful LLM to effectively fuse geometric, appearance, and linguistic information. To overcome the scarcity of point-text instruction following data, we developed an automated data generation pipeline, collecting a large-scale dataset of more than 730K samples with 660K different objects, which facilitates the adoption of the two-stage training strategy prevalent in MLLM development. Additionally, we address the absence of appropriate benchmarks and the limitations of current evaluation metrics by proposing two novel benchmarks: Generative 3D Object Classification and 3D Object Captioning, which are supported by new, comprehensive evaluation metrics derived from human and GPT analyses. Through exploring various training strategies, we develop PointLLM, significantly surpassing 2D and 3D baselines, with a notable achievement in human-evaluated object captioning tasks where it surpasses human annotators in over 50\% of the samples. Codes, datasets, and benchmarks are available at \href{https://github.com/OpenRobotLab/PointLLM}{https://github.com/OpenRobotLab/PointLLM}.
\keywords{Multi-Modal LLM \and 3D Understanding \and Point Cloud}
\end{abstract}

\begin{figure}[t]
\centering
    \includegraphics[width=\linewidth]{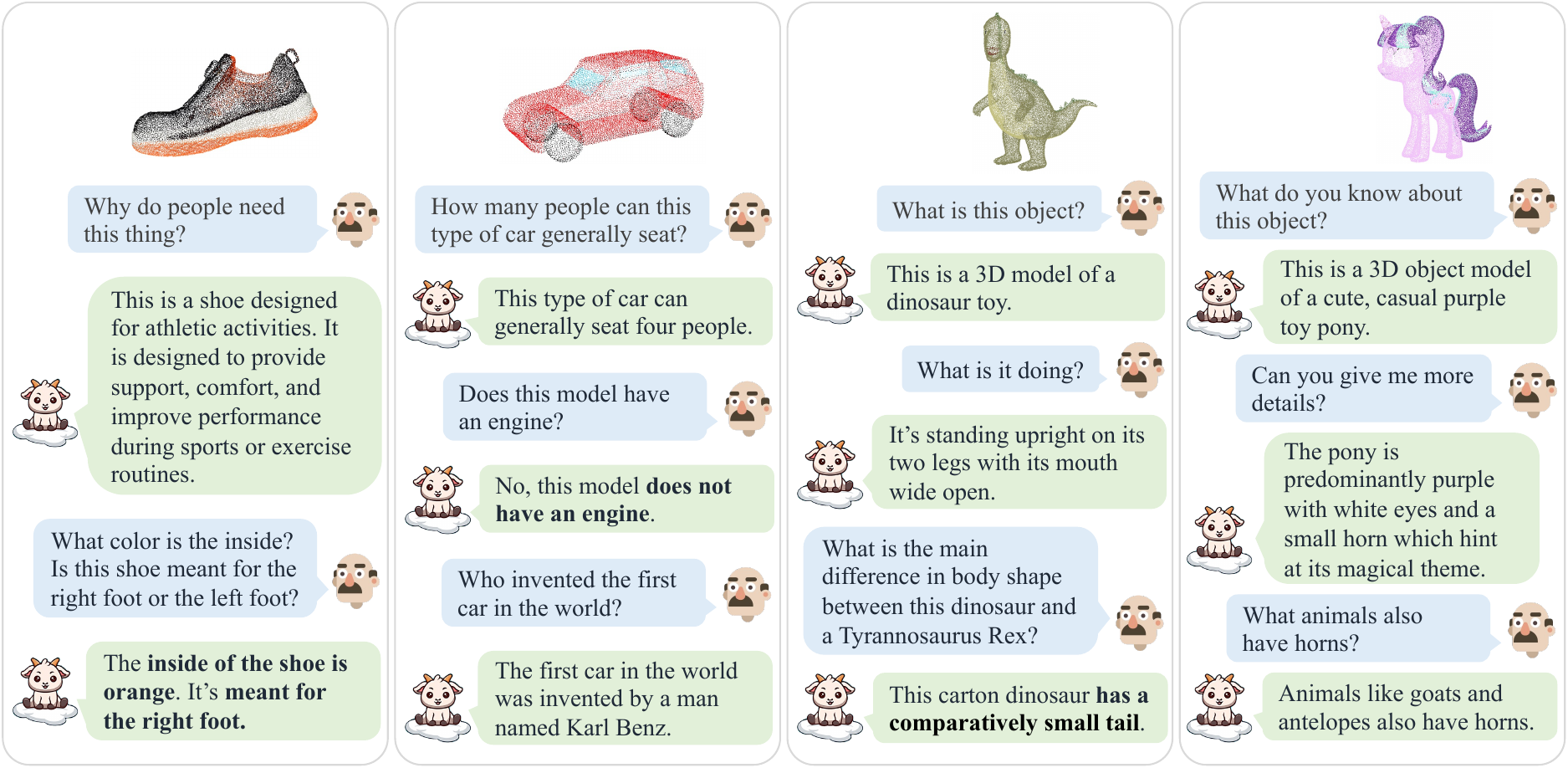}
\caption{We introduce PointLLM, a multi-modal large language model capable of understanding colored point clouds of objects. It perceives object types, geometric, and appearance without concerns for ambiguous depth, occlusion, or viewpoint dependency.}
\label{fig:teaser}
\end{figure}

\section{Introduction}

Recent developments in large language models (LLMs) \cite{ChatGPT,GPT-3,GPT-4,InstructGPT,InternLM,Palm,LLaMA,T5} have showcased their remarkable capabilities in natural language processing, acting as generalized interfaces \cite{GeneralInterface} for a broad range of tasks\cite{T5,GPT-3}. Beyond text, the exploration of multi-modal LLMs (MLLMs) now extends to processing audio\cite{AudioGPT}, images\cite{LLaMA-Adapter, LLaVA, MiniGPT-4, Kosmos-1, GPT-4, Flamingo, BLIP-2}, and more. 

The next step in this evolution lies in understanding 3D structures, and particularly point clouds. Suppose we want to embed LLMs in 3D design software for interactive 3D content creation/editing via text commands (\ie 3D copilot), this requires LLMs to understand 3D content states, which can be represented as point clouds. In robotics, LLMs as control centers need to understand the environments for perception and planning, where point clouds captured through depth sensors or LiDAR are important observations. 

While existing efforts to integrate LLMs with 2D images\cite{LLaVA,InstructBLIP,MiniGPT-4,Palm-E} also provide 3D comprehension, they face difficulties like depth ambiguity, occlusion, and viewpoint dependency. In contrast, point clouds, as an efficient and universal 3D representation, offer direct geometric and appearance information. Despite these benefits, the integration of point clouds with LLMs is still underexplored.


Recently, connecting pre-trained encoders with LLMs using projection layers and employing a two-stage training of alignment and instruction tuning has been proven effective for developing MLLMs across various domains\cite{LLaVA, Palm-E, InstructBLIP, MiniGPT-4, Kosmos-1, MotionGPT}. We pose a question: Can this established framework be successfully adapted to the realm of point clouds? In this work, we affirmatively answer this question by introducing PointLLM (\cref{fig:teaser}), our preliminary effort to empower LLMs to understand point clouds, with a focus on 3D objects.


The first difficulty to be address is the lack of training data, the point-text instruction following datasets essential for teaching models to interpret point clouds and follow user commands. While manual compilation is costly and labor-intensive, we devised an automated data collection pipeline using GPT-4 to generate diverse instructions from Objaverse's\cite{Objaverse} captions. This produced a large-scale dataset comprising 660K brief-description for different objects and 70K complex instructions, enabling the model's two-stage training for this domain.

Evaluating model performance with appropriate tasks and metrics presents another challenge. We aim to assess point cloud comprehension in MLLMs and existing discriminative-based 3D perception benchmarks fall short for this purpose due to the generative nature of MLLMs. We introduce two novel benchmarks: Generative 3D Object Classification and 3D Object Captioning, based on a hypothesis that LLMs’ understanding of point clouds is reflected by their ability to identify the object’s category and the accuracy and details of captions, which elaborate the information they perceive. We also observe the limitations of some widely used NLP metrics like BLEU-1\cite{BLEU}, ROUGE-L\cite{ROUGE}, and METEOR\cite{METEOR} for their short caption bias and inability to reflect linguistic diversity. To counter these shortcomings, we devise new evaluation metrics that combine human and GPT-4/ChatGPT evaluation with data-driven approaches, establishing a comprehensive evaluation framework. To our knowledge, we are the first to introduce generative object classification in this field.


In our study, we evaluated various training recipes and observed that an optimal number of projection layers enhances feature clustering, aligning point and text features effectively. We also found that employing max pooling to aggregate point tokens reduces the token number and greatly enhances training speed, albeit with a slight performance trade-off. Further analysis of data variability revealed that the model performance peaks with about 600K samples for alignment and diverse instruction data notably enhances fine-tuning, emphasizing the value of data quantity and diversity. These insights led to the development of PointLLM, which markedly surpasses 2D and 3D baselines, impressively scoring higher than human annotators in over 50\% of the object captioning samples.

\section{Related Work}

\noindent\textbf{Multi-modal large language models.}
Multi-modal Large Language Models (MLLMs) are designed to comprehend and interpret a wide range of information that extends beyond mere text-based data\cite{MLLM-survey}, including but not limited to images\cite{Kosmos-1, LLaVA, MMGPT, MiniGPT-4, VisionLLM}, audio\cite{AudioGPT}, motion\cite{MotionGPT}, etc. Broadly, the models can be classified into two categories. The first category includes models that employ a large language model to interface with individual, modality-specific models or APIs \cite{VisualChatGPT, ViperGPT, VisualProgramming, AudioGPT, Gorilla}. This approach circumvents the need for model training but is heavily dependent on the availability and capabilities of pre-existing models or APIs. The second category pertains to models that employ an end-to-end training strategy. There are two prominent paradigms within this category. The first involves training the model from scratch, similar to text-only LLMs, using large-scale multi-modal corpora and datasets \cite{Kosmos-1, Kosmos-2}. The second paradigm builds on pre-trained LLMs and unimodal encoders.\cite{Palm-E, BLIP-2, Flamingo, OpenFlamingo, LLaMA-Adapter, LLaMA-Adapterv2, LLaVA, MMGPT, MiniGPT-4, Otter, InstructBLIP, PanadaGPT, GPT4RoI}. This strategy typically involves a two-stage process: alignment of the unimodal encoder with the LLM's feature space, followed by instruction-based fine-tuning. In our work, we adhere to the alignment and tuning strategy to construct an MLLM capable of understanding 3D object point clouds.

\noindent\textbf{Object point cloud understanding with language.}
Inspired by models like CLIP \cite{CLIP}, which bridges visual and textual modalities, similar advancements have emerged in the 3D object domain\cite{PointCLIP,PointCLIPv2,CLIP2Point,JM3D,OpenShape,ULIP,ULIP-2,CG3D}. PointCLIP\cite{PointCLIP}, PointCLIPv2\cite{PointCLIPv2}, and CLIP2Point\cite{CLIP2Point} utilize depth image projections of point clouds for 3D recognition with 2D CLIP models. Others, such as ULIP\cite{ULIP}, JM3D\cite{JM3D}, OpenShape\cite{OpenShape}, and CG3D\cite{CG3D}, train point cloud encoders to align with CLIP representations using triplets of point clouds, images, and texts. ULIP-2\cite{ULIP-2} and OpenShape\cite{OpenShape} have expanded this by employing image-captioning models for automatic data generation, enhancing training triplet scalability. Cap3D \cite{Cap3D} and UniG3D \cite{UniG3D} adopt similar approaches for point-text dataset generation. In our work, we leverage Cap3D's captions on Objaverse for automatic instruction-data generation in training PointLLM. The recently introduced 3D-LLM\cite{3D-LLM} also seeks to enable LLMs to comprehend 3D, by rendering objects into multi-view images, using 2D foundational models like CLIP\cite{CLIP} and SAM\cite{SAM} for feature extraction, and 2D MLLMs such as BLIP\cite{BLIP-2} for output generation. Concurrently, Point-Bind LLM\cite{Point-Bind} aligns point cloud features with ImageBind\cite{ImageBind} and uses 2D MLLMs like Imagebind-LLM\cite{ImageBind-LLM} for generation. Though simple, it faces challenges like hallucination due to its retrieval nature. Distinctively, PointLLM directly aligns point clouds with LLM by end-to-end training, avoiding complicated data pre-processing and enabling accurate, open-ended, and free-form interactions.
\section{Methodology}

This section elucidates our strategy for the automatic generation of point-text instruction-following data. We then delve into the architecture of PointLLM, which takes as input an object point cloud and user instruction and outputs responses. Lastly, we detail our loss function and two-stage training strategy.

\subsection{Point-Text Instruction Following Data}
The daunting challenge in the development of an end-to-end multi-modal LLM is procuring large-scale multi-modal instruction-following data, vital for representation learning, aligning latent spaces, and orienting the model to adhere to human intentions\cite{Flamingo, BLIP, InstructBLIP, LLaVA, MiniGPT-4}. However, manual labeling of such data is cost-prohibitive and labor-intensive. 
To overcome this, we follow \cite{LLaVA} and propose an automated data generation technique utilizing the large-scale point cloud captioning dataset, Cap3D\cite{Cap3D}, with the assistance of GPT-4\cite{GPT-4}. The generated dataset adheres to a uniform instruction following template, shown in \cref{tab:instruction_following_template}, and consists of brief-description instructions and complex instructions, which aid in latent space alignment and instruction tuning, respectively.
\begin{table}[t!]
\centering
\caption{\textbf{Instruction following template.} \textcolor{bluecolor}{\{System Prompt\}} is the system prompt used by the pre-trained LLM, \textcolor{bluecolor}{\{p\_tokens\}} are point tokens, and \textcolor{bluecolor}{\{Instruction\}} and \textcolor{greencolor}{\{Response\}} denote user instructions and model responses. Losses are computed only for model responses and the end-of-sentence token \texttt{<}/{}s\texttt{>}.}

\label{tab:instruction_following_template}
\scalebox{0.89}{\tablestyle{8pt}{1.0}
\begin{tabular}{@{}l@{\hspace{0.2em}} p{0.55\linewidth}@{}} 
\toprule
\textcolor{bluecolor}{\texttt{\{}System Prompt\texttt{\}}} & \\ \midrule
USER: & 
\begin{minipage}[t]{\linewidth}
\texttt{<}p\_start\texttt{>}\textcolor{bluecolor}{\texttt{\{}p\_tokens\texttt{\}}}\texttt{<}p\_end\texttt{>}\textcolor{bluecolor}{\{Instruction 1\}}
\end{minipage} \\ 
ASSISTANT: & \textcolor{greencolor}{\{Response 1\}\texttt{<}/{}s\texttt{>}} \\ \midrule
USER: & \textcolor{bluecolor}{\{Instruction 2\}} \\
ASSISTANT: & \textcolor{greencolor}{\{Response 2\}\texttt{<}/{}s\texttt{>}} \\ \midrule
USER: & \textcolor{bluecolor}{\{Instruction 3\}} \\
ASSISTANT: & \textcolor{greencolor}{\{Response 3\}\texttt{<}/{}s\texttt{>}} \\
\bottomrule
\end{tabular}
}
\end{table}

\noindent\textbf{Brief-description instructions.}
The Cap3D\cite{Cap3D} dataset provides two variations of captions for the 3D objects in Objaverse\cite{Objaverse}: those generated by image-captioning models and those annotated by humans. While there are 660K objects accompanied by generated captions, only 40K samples have human-annotated captions. For brief-description instruction, we utilize the model-generated split due to the need for a larger data volume for aligning the latent spaces of point cloud and text modalities \cite{LLaVA}. We created a list of 30 instructions to instruct the model to provide a succinct description of a given 3D object point cloud. A random instruction from this list is chosen as the user instruction, and the caption from Cap3D is used directly as the model response, forming a single-round instruction following sample. This results in 660K brief-description instruction data, each corresponding to a unique object point cloud.

\noindent\textbf{Complex instructions.}
Beyond brief descriptions, it's crucial that the model learns to understand objects from a variety of angles, responding accurately to diverse human instructions. To facilitate this, we employ GPT-4 to produce complex instruction-following data. Specifically, a caption from Cap3D is used to stimulate GPT-4 to craft a more comprehensive description that identifies the object's type, appearance, functionalities, and any other inferable information. Similar to the process for generating brief-description instructions, we also curate a set of 30 distinct prompts, each pushing the model to describe the 3D object in depth. One of these prompts is randomly coupled with the newly crafted description, forming a training sample. GPT-4 is further used to generate conversations (\ie, Q\&A pairs) that delve into diverse aspects of the object based on the captions, such as the object's functionality or materials, and the corresponding answers should be informative and comprehensive. For each object, GPT-4 generates 3 single-round conversations and 1 multi-round conversation with 3 Q\&A pairs, all ensuring logical relevance.

\begin{figure}[ht]
\centering
\includegraphics[width=\linewidth]{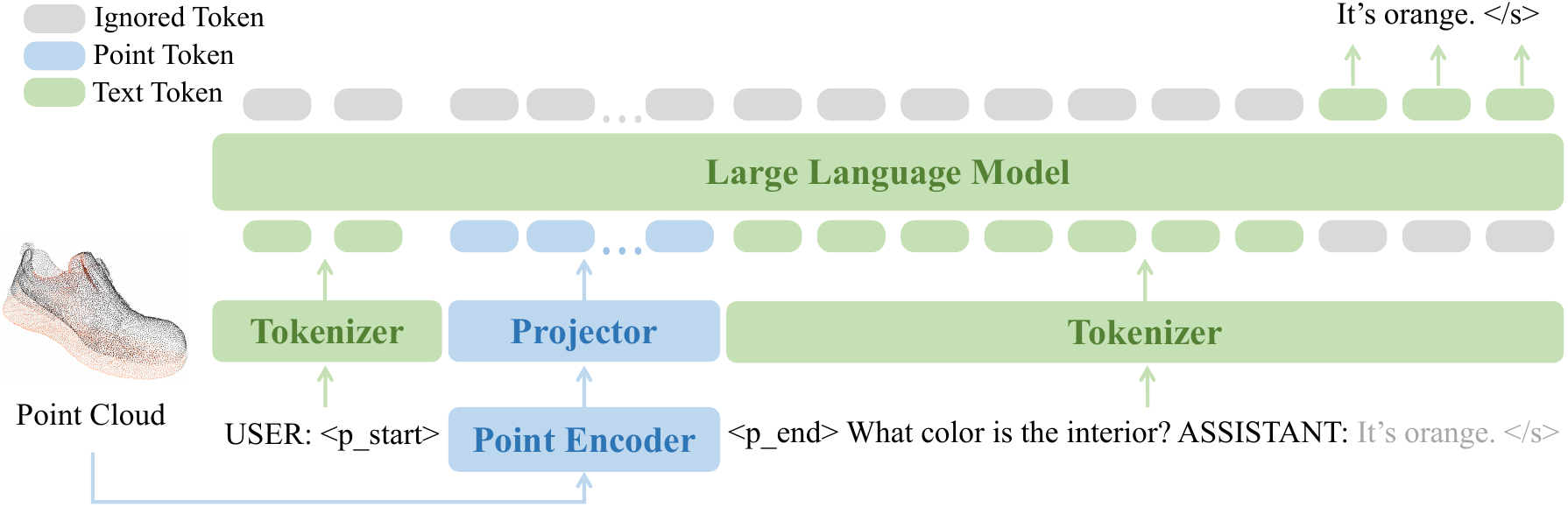}
\caption{\textbf{An overview of PointLLM.} The point encoder extracts features from the input point cloud and the projector projects them to the latent space of the LLM backbone. The LLM backbone processes sequences of point and text tokens and generates the predicted tokens as the output.}
\label{fig:main_figure}
\end{figure}

With a focus on data quality, we selected 15K captions from the Cap3D human-annotated split for data generation, each comprising more than five words. After filtering incorrect GPT-4 outputs, we collected 70K complex instructions, including 15K detailed descriptions, 40K single-round conversations, and 15K multi-round conversations. The instruction lists, GPT-4 prompts, data examples, and distribution analysis can be found in the supplementary material.

\subsection{Model Architecture}
As shown in \cref{fig:main_figure}, our PointLLM is a generative model that aims to complete multi-modal sentences containing both point clouds and texts. The model consists of three main components: a pre-trained point cloud encoder $f_{pe}$, a projector $f_{proj}$, and a pre-trained large language model (LLM) backbone $f_{llm}$. 

The point cloud encoder $f_{pe}$ takes as input a point cloud $P \in \mathbb{R}^{n \times d}$, where $n$ is the number of points and $d$ is the feature dimension of each point. The output of the encoder is a sequence of point features $X = (x_1, x_2, …, x_m) \in \mathbb{R}^{m \times c}$, where $m$ is the number of point features and $c$ is the feature dimension. The projector $f_{proj}$ is a MLP that maps the point features $X$ to point tokens $Y = (y_1, y_2, ..., y_m) \in \mathbb{R}^{m \times c'}$, where $c'$ is the dimension of the point tokens, which is the same as the text tokens.

The LLM backbone $f_{llm}$ is a decoder-only Transformers~\cite{Attention}, which accepts a sequence of tokens, composed of both text and point tokens. This mixed sequence of tokens is denoted as $Z = (z_1, z_2, ..., z_k) \in \mathbb{R}^{k \times c'}$, where $k$ is the total number of tokens. Utilizing a self-attention mechanism, the LLM backbone is capable of understanding the contextual relationships between different types of tokens, enabling it to generate responses based on both text and point cloud inputs. 
Formally, the output of the LLM backbone $f_{llm}$ is a sequence of predicted tokens $\hat{Z} = (\hat{z}_1, \hat{z}_2, ..., \hat{z}_k) \in \mathbb{R}^{k \times c'}$. The prediction of the $i$-th token, $\hat{z}_i$, is conditioned on all previous tokens, $Z_{<i} = (z_1, ..., z_{i-1})$, expressed mathematically as 
\begin{equation}
    \hat{z}_i = f_{llm}(Z_{<i}).
\end{equation}
Each $\hat{z}_i$ is passed through a final linear layer followed by a softmax operation, mapping the hidden states into a probability distribution over the vocabulary. This additional layer is denoted as $f_{vocab}: \mathbb{R}^{c'} \to \mathbb{R}^{V}$, where $V$ is the size of the vocabulary. The final prediction $\tilde{z}_i$ for the $i$-th token is the word in the vocabulary with the highest probability: 
\begin{equation}
    \tilde{z}_i = \arg\max_{w \in \text{vocab}} f_{vocab}(\hat{z}_i)[w].
\end{equation}
\subsection{Training}

\noindent\textbf{Loss function.}
We train PointLLM by minimizing the negative log-likelihood of the text token at each position. Our loss function is only computed on text tokens that constitute the model's responses, including the end-of-sentence token \texttt{<}/{}s\texttt{>}. We exclude the tokens from human instructions, ensuring that the model focuses on learning to generate accurate and coherent responses. The end-to-end nature of this training approach enables PointLLM to effectively integrate point cloud and text modalities.

\noindent\textbf{Two-stage training.}
Our training procedure comprises two stages, each focusing on different aspects of the model.

During the first stage, termed the \textbf{feature alignment stage}, we freeze the parameters of the point cloud encoder and the LLM, and train only the MLP projector. At this stage, the training process uses brief-description instructions, aiming to align point features with the text token space effectively. This stage also includes the adjustment of token embeddings for the two newly added special tokens \texttt{<}p\_start\texttt{>} and \texttt{<}p\_end\texttt{>}.

In the second stage, referred to as the \textbf{instruction tuning stage}, we freeze the point cloud encoder while jointly training the projector and the LLM. This second stage uses complex instructions and helps the model build its ability to understand and respond to complex instructions including point cloud data.
\section{Benchmarks and Evaluation}

Evaluating multi-modal LLMs presents a challenge due to the lack of a unified metric for their diverse outputs. Current 3D point cloud benchmarks primarily focus on discriminative tasks, such as classification or retrieval, missing the generative and open-vocabulary aspects of LLMs. To address this, we introduce two novel benchmarks: Generative 3D Object Classification and 3D Object Captioning, designed to assess model generalization and understanding of point clouds. Our underlying hypothesis is that models' comprehension of point clouds, at the very least, manifests in the accurate classification of objects. Furthermore, this comprehension is proportional to the accuracy and details of the captions, which elaborate the information they perceive. To support these benchmarks, we've developed a novel and comprehensive evaluation framework that incorporates human, GPT-4/ChatGPT, and traditional metrics. The supplementary material provides the prompts and the human verification of the GPT evaluation.

\subsection{Generative 3D Object Classification}
The task of generative 3D object classification involves prompting the model to freely answer the object type from its point cloud, distinguishing it from discriminative models that classify objects based on probability comparisons. We consider two settings: close-set zero-shot and open-vocabulary setting.

\noindent\textbf{Close-set zero-shot classification.} In this scenario, the object type belongs to a fixed set of categories, and the model never sees any samples of this dataset during training. This tests the model’s ability to generalize to unseen domains. We use the test split of the ModelNet40 \cite{ModelNet40} dataset as our source of data, which contains point clouds of 40 different object categories. We use ChatGPT as a post-processor to select one of the ModelNet40 categories based on the model’s answer. If ChatGPT selects the correct option, then we consider the model’s classification correct; otherwise, we consider it incorrect. Please refer to the supplementary material for more discussions about this task's setting.

\noindent\textbf{Open-vocabulary classification.} In this scenario, the object type is not limited to a predefined set of categories, but can be any word or phrase that identifies the object. This reflects the real-world setting where new objects can appear at any time, and the model needs to be able to recognize them without retraining. We randomly select 200 objects from the Objaverse\cite{Objaverse} dataset, incorporating human-annotated captions from Cap3D\cite{Cap3D} as ground truth labels for the task. We employ GPT-4 to verify if the model's response matches the intended object type as described in the human caption, allowing for varied expressions that correctly identify the object. For instance, responses like "a cup" or "a coffee mug" are considered correct classification for a human caption of "a blue mug." GPT-4 is preferred in this setting for its precision in recognizing synonymous object descriptions, in contrast to ChatGPT, which is more prone to false negatives by not acknowledging similar terms for the same object. ChatGPT, however, is used in the close-set setting as it performs accurately but at 95\% less cost.

\subsection{3D Object Captioning}
3D object captioning evaluates the model's detailed understanding of point clouds. We utilize the same 200 objects as for the open-vocabulary classification and prompt our model to caption them in detail. Human-annotated captions serve as reference ground truths for automatic evaluation.

For a comprehensive and robust evaluation, we employ three distinct methods to assess performance in this task:
\begin{enumerate}
  \item \textbf{Human evaluation.} Human evaluators conduct a binary classification and counting task, reviewing randomly shuffled captions from various models and human-annotated captions for the objects, without knowing their sources. Using the official Objaverse\cite{Objaverse} explorer, evaluators visually inspect each object and systematically assess each object attribute (such as type, color, material, \etc) mentioned in the model's response, assigning one correct or hallucination point for each attribute based on its accuracy. The aggregate of these points forms the correctness and hallucination scores. Precision, the ratio of accurate information in the model's response, is then calculated. For detailed scoring criteria, please refer to the supplementary material.
  \item \textbf{GPT-4 evaluation.} Given a model-generated caption and its corresponding human reference, GPT-4 identifies the attributes mentioned in the human caption and calculates the percentage of these attributes that are either correctly or partially matched in the model's caption, scoring from 0 to 100 with an explanation. As one of the most advanced language models, GPT-4 is well-equipped to perform such tasks. Our method of calculating the correct percentage and assigning scores offers an advantage over approaches like those in \cite{LLaVA}, which directly generate an overall score without transparency.
  \item \textbf{Traditional metric evaluation.} We report traditional metrics results including BLEU-1 \cite{BLEU}, ROUGE-L \cite{ROUGE}, and METEOR \cite{METEOR} following \cite{3D-LLM}. Though widely used, these metrics have limitations as detailed in \cref{sec:limitation_traiditional}. Therefore, we incorporate and rely more on two additional data-driven metrics, Sentence-BERT \cite{Sentence-BERT} and SimCSE\cite{SimCSE} similarity, which compute the similarity of sentence embeddings between model-generated and human captions.
\end{enumerate}
\section{Experimental Results}
\label{sec:exp}

\subsection{Experimental Settings}

\noindent\textbf{Implementation details.} We use the LLaMA model\cite{LLaMA} as our LLM backbone, with the 7B and 13B Vicuna\cite{Vicuna} checkpoint. Point-BERT\cite{Point-BERT}, pre-trained with ULIP-2\cite{ULIP-2} on the Objaverse \cite{Objaverse} dataset, serves as our point encoder. The 200 objects from Objaverse utilized for our benchmarks are not seen during any stage of the training. We utilize $n=8192$ points and $d=6$ dimensions for each point cloud. We assign a black color to point clouds from ModelNet40, as they lack color information. The point encoder outputs $m=513$ point features, each with $c=384$ dimensions. The projector contains three linear layers with the GeLU\cite{GeLU} activation, which maps point features to tokens with $c'=5120$ (7B model) or $c'=5120$ (13B model) dimensions. As we add two additional special tokens, the vocabulary size of PointLLM is $V=32003$. All experiments are conducted on 8 $\times$ 80G A100 GPUs. GPT-4 and ChatGPT in this paper all refer to OpenAI's "gpt-4-0613" and "gpt-3.5-turbo-0613" models respectively. More implementation and training details are provided in the supplementary material.

\noindent\textbf{Baselines.} Our analysis includes comparisons with baselines capable of performing the same generative classification and captioning tasks. We focus on 3D-LLM\cite{3D-LLM} and Point-Bind LLM\cite{Point-Bind}; 3D-LLM is assessed solely on the Objaverse dataset due to its current lack of support for pure point clouds, while Point-Bind LLM, not supporting colored point clouds, is excluded from captioning. We also compare with two popular 2D MLLMs, InstructBLIP\cite{InstructBLIP} and LLaVA\cite{LLaVA}, to explore the performance gap between image-based and point-based MLLMs and to highlight the advantages of point clouds over single-view~images.

\begin{table}[t!]
\centering
\caption{\textbf{Generative 3D object classification results on the ModelNet40 (M40.) test split and Objaverse (Obj.).} The results show the classification accuracy under the \textbf{I}nstruction-typed (I) prompt "What is this?" and the \textbf{C}ompletion-typed (C) prompt "This is an object of " as well as the average accuracy.}
\label{tab:generative_classification}
\scalebox{0.94}{\tablestyle{2pt}{1.0}

\begin{tabular}{l@{\hspace{-0.2em}}c@{\hspace{-0.3em}}ccccc}
\toprule
Model & Input & M40.(I) & M40.(C) & Obj.(I) & Obj.(C) & Avg. \\ \midrule
InstructBLIP-7B\cite{InstructBLIP} & Single-V. Img. & 19.53 & 31.48 & 45.00 & 42.00 & 34.50 \\
InstructBLIP-13B\cite{InstructBLIP} & Single-V. Img. & 25.97 & 31.40 & 37.00 & 31.50 & 31.47 \\
LLaVA-7B\cite{LLaVA} & Single-V. Img. & 39.75 & 39.67 & 49.50 & 50.50 & 44.86 \\
LLaVA-13B\cite{LLaVA} & Single-V. Img. & 37.12 & 36.06 & 53.00 & 50.50 & 44.17 \\ \midrule
3D-LLM\cite{3D-LLM} & 3D Obj. + Mul.-V. Img. & - & - & 49.00 & 41.50 & 45.25 \\
Point-Bind LLM\cite{Point-Bind} & 3D Point Cloud & 51.90 & 39.71 & 6.00 & 4.50 & 25.53 \\
\textbf{PointLLM-7B} & 3D Point Cloud & \textbf{53.44} & 51.82 & 55.00 & 51.00 & 52.82 \\
\textbf{PointLLM-13B} & 3D Point Cloud & 53.00 & \textbf{52.55} & \textbf{56.50} & \textbf{51.50} & \textbf{53.39} \\ \bottomrule
\end{tabular}
}
\end{table}

\subsection{Generative 3D Object Classification}
\cref{tab:generative_classification} shows the classification accuracy of various models on our proposed tasks. For 2D MLLMs' image inputs, we randomly sample rendered images of ModelNet40 point clouds and Objaverse objects. We prompt all the models with the same prompts of two types: the \textbf{I}nstruction-typed (I) prompt "What is this?" and the \textbf{C}ompletion-type (C) prompt "This is an object of~".

Experimental results demonstrate PointLLM's superiority over both 2D and 3D MLLMs on ModelNet40 and Objaverse datasets for various prompt types. Compared with 2D models, PointLLM offers direct point cloud engagement, showcasing enhanced 3D object comprehension over single-view images. This method effectively addresses challenges like occlusion and viewpoint variation, leveraging rich 3D geometry and appearance data from colored point clouds. PointLLM shows more consistent classification accuracy across different prompts than other 3D models, underlining its prompt robustness. Utilizing a pre-trained point encoder and an LLM backbone, PointLLM efficiently translates point cloud data into descriptive natural language, capturing the object's identity.

The zero-shot performance on ModelNet40 further illustrates our model's aptitude for generalization. Even though ModelNet40 comprises point clouds unseen during training, PointLLM recognizes them using its pre-existing knowledge and perception abilities honed during our two-stage training. This adaptability to unseen domains and novel objects, without necessitating retraining, is crucial for real-world deployment as a foundation model.

\subsection{3D Object Captioning}
\label{exp:3d_captioning}
\cref{tab:object_captioning} displays the results of the captioning benchmark, averaged across objects. Each model was prompted with "Caption this 3D model in detail."

\begin{table}[t!]
\centering
\caption{\textbf{3D object captioning results on Objaverse.} Evaluation encompasses human (correctness, hallucination, precision) and GPT-4 assessments, supplemented by Sentence-BERT, SimCSE, BLEU-1, ROUGE-L, and METEOR metrics. A primary focus is placed on human and GPT-4 evaluation, along with data-driven metrics. "*" indicates PointLLM was prompted for shorter captions with no more than 20 words.}
\label{tab:object_captioning}
\scalebox{0.93}{\tablestyle{1.8pt}{1.0}

\begin{tabular}{lcccccc
>{\columncolor[HTML]{EFEFEF}}c 
>{\columncolor[HTML]{EFEFEF}}c 
>{\columncolor[HTML]{EFEFEF}}c}
\toprule
Model & Corr. & Hallu.$\downarrow$ & Prec. & GPT-4 & S.-BERT & SimCSE & B-1. & R-L. & MET. \\ \midrule
InstructBLIP-7B\cite{InstructBLIP} & 2.56 & 0.77 & 76.99 & 45.34 & 47.41 & 48.48 & 4.27 & 8.28 & 12.99 \\
InstructBLIP-13B\cite{InstructBLIP} & 2.58 & 1.13 & 69.56 & 44.97 & 45.90 & 48.86 & 4.65 & 8.85 & 13.23 \\
LLaVA-7B\cite{LLaVA} & 2.76 & 0.86 & 76.30 & 46.71 & 45.61 & 47.10 & 3.64 & 7.70 & 12.14 \\
LLaVA-13B\cite{LLaVA} & 2.43 & 0.86 & 73.97 & 38.28 & 46.37 & 45.90 & 4.02 & 8.15 & 12.58 \\ \midrule
3D-LLM\cite{3D-LLM} & 1.77 & 1.16 & 60.39 & 33.42 & 44.48 & 43.68 & 16.91 & 19.48 & \textbf{19.73} \\
\textbf{PointLLM-7B} & 3.04 & \textbf{0.66} & \textbf{82.14} & 44.85 & 47.47 & 48.55 & 3.87 & 7.30 & 11.92 \\
\textbf{PointLLM-13B} & \textbf{3.10} & 0.84 & 78.75 & \textbf{48.15} & \textbf{47.91} & \textbf{49.12} & 3.83 & 7.23 & 12.26 \\ 
\rowcolor[HTML]{EFEFEF}\textbf{PointLLM-13B*} & 2.12 & 0.39 & 84.39 & 44.27 & 50.15 & 50.83 & \textbf{17.09} & \textbf{20.99} & 16.45 \\ \midrule
Human & 2.67 & 0.22 & 92.46 & 100.00 & 100.00 & 100.00 & 100.00 & 100.00 & 100.00 \\ \bottomrule
\end{tabular}}
\end{table}

In \cref{tab:object_captioning} our models significantly outperform all baselines in key evaluation metrics for 3D object captioning, especially in human correctness score and GPT-4 evaluations. 
These scores reflect a model's ability to capture and articulate the intricate details of objects. Notably, PointLLM achieves the highest correctness scores, producing more accurate and detailed captions than other models, even rivaling human annotations. In addressing hallucination, a common challenge, our PointLLM exhibits the lowest hallucination scores and highest precision scores, indicating its effectiveness in generating detailed, accurate captions with less false information. The Sentence-BERT and SimCSE results further confirm our model's capability in producing captions more semantically aligned with the ground truth. Interestingly, all 13B models, regardless of being 2D or 3D MLLMs, 
\begin{wrapfigure}{r}{0.54\textwidth}
  \centering
    \includegraphics[width=0.53\textwidth]{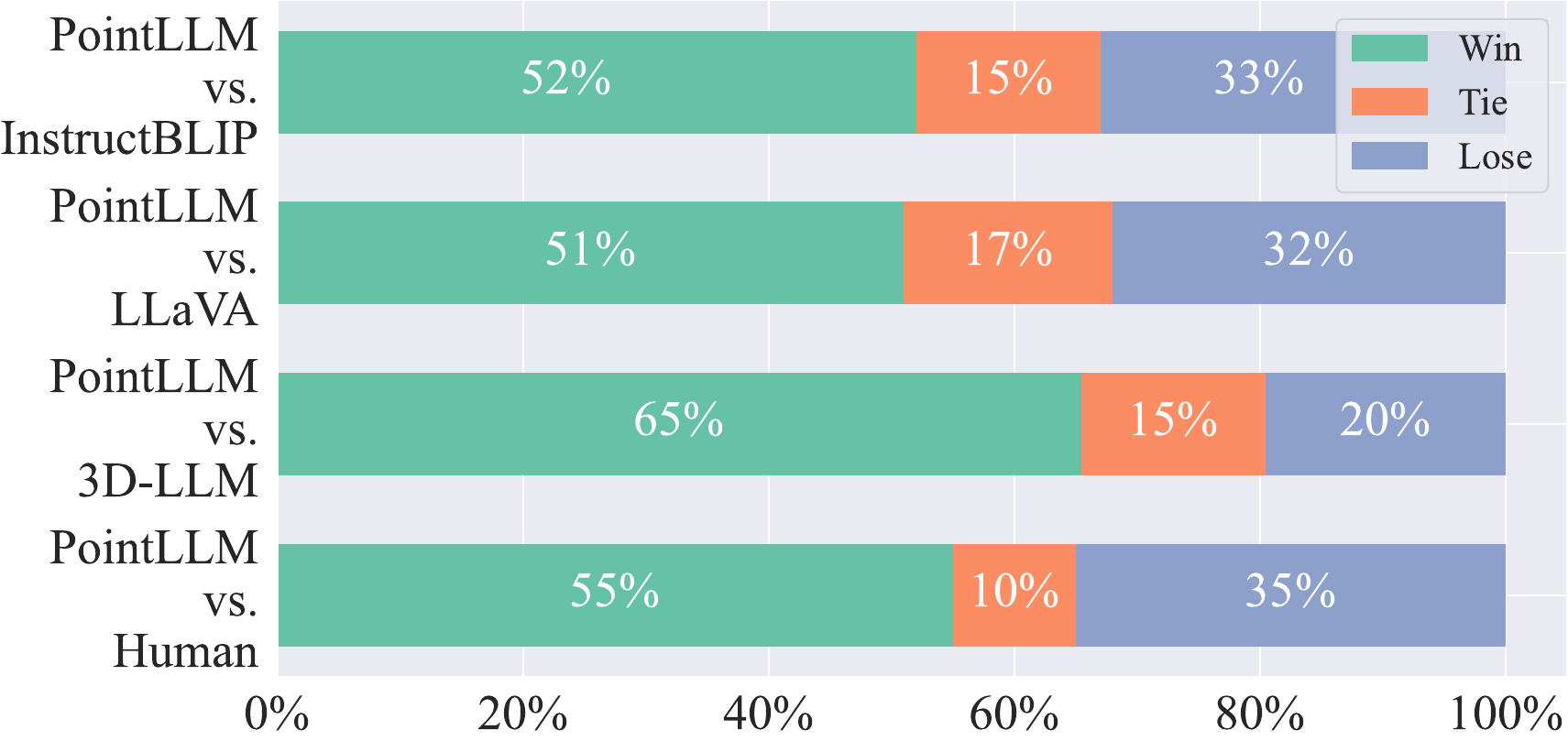}
  \caption{\textbf{Win rate comparison}.}
  \label{fig:win-rate}
\end{wrapfigure}
have higher hallucination scores than their 7B counterparts. This suggests larger MLLMs may be more challenging to fine-tune for precision. The investigation of this trend and its causes is an intriguing research direction.

We analyzed the human evaluation data to compare our models with baselines
and human annotations. Win rates, calculated based on the correctness score for the 13B variants, are presented in \cref{fig:win-rate}. PointLLM demonstrates notable performance, outperforming counterparts in over half of the test samples, including against human annotations (55\% vs. 35\%). This underscores PointLLM's ability to effectively capture and convey 3D object details, hinting at its potential for scalable, human-like captioning of 3D objects. More win rate comparisons are detailed in the supplementary material.

\label{sec:limitation_traiditional}
\noindent\textbf{Limitations of traditional metrics.} Our evaluation also highlights the limitations of conventional NLP metrics like BLEU-1, ROUGE-L, and METEOR. These metrics are biased toward shorter captions. For instance, 3D-LLM achieves higher scores on these metrics by producing shorter captions (averaging 20 words compared to others' 69+) that do not necessarily indicate superior quality, as confirmed by human evaluation. To further verify this, we prompt PointLLM-13B to generate captions with no more than 20 words, which improved these metric scores significantly, as shown in \cref{tab:object_captioning}. However, this preference for short captions contradicts our benchmark, which necessitates that models produce detailed captions to demonstrate a comprehensive understanding of point clouds. Also, these metrics often fail to capture the semantic similarity or diversity of the captions as they primarily measure the overlap of n-grams or their varieties. An example in \cref{tab:traditional_metrics} highlights this: inaccurately describing a "Private jet" as a "jet engine" scores higher compared to accurately identifying it as an "airplane". It's worth noting that these metrics are proposed for machine translation, not captioning. 
\begin{table}[t!]
\centering
\caption{\textbf{Biased Traditional metrics for different captions.} The biased scores demonstrate the limitations of these metrics.}
\label{tab:traditional_metrics}
\scalebox{0.94}{\tablestyle{8pt}{1.0}
\begin{tabular}{@{}p{9cm}@{\hspace{-1.5em}}c@{\hspace{0.8em}}c@{\hspace{0.8em}}c@{}}
\toprule
Caption                 & B-1. & R-L. & MET. \\ \midrule
\textbf{Private jet}             & \textbf{100.00} & \textbf{100.00}  & \textbf{100.00} \\
there is a black jet engine in a dark background & 10.00 & 18.18 & 17.86 \\
This is a 3D model of a cartoon-style commercial airplane. & 0.00   & 0.00  & 0.00   \\ \bottomrule
\end{tabular}
}
\end{table}
\begin{figure}[t]
    \centering
    \begin{minipage}{0.44\textwidth}
        \centering
        \includegraphics[width=0.6\linewidth]{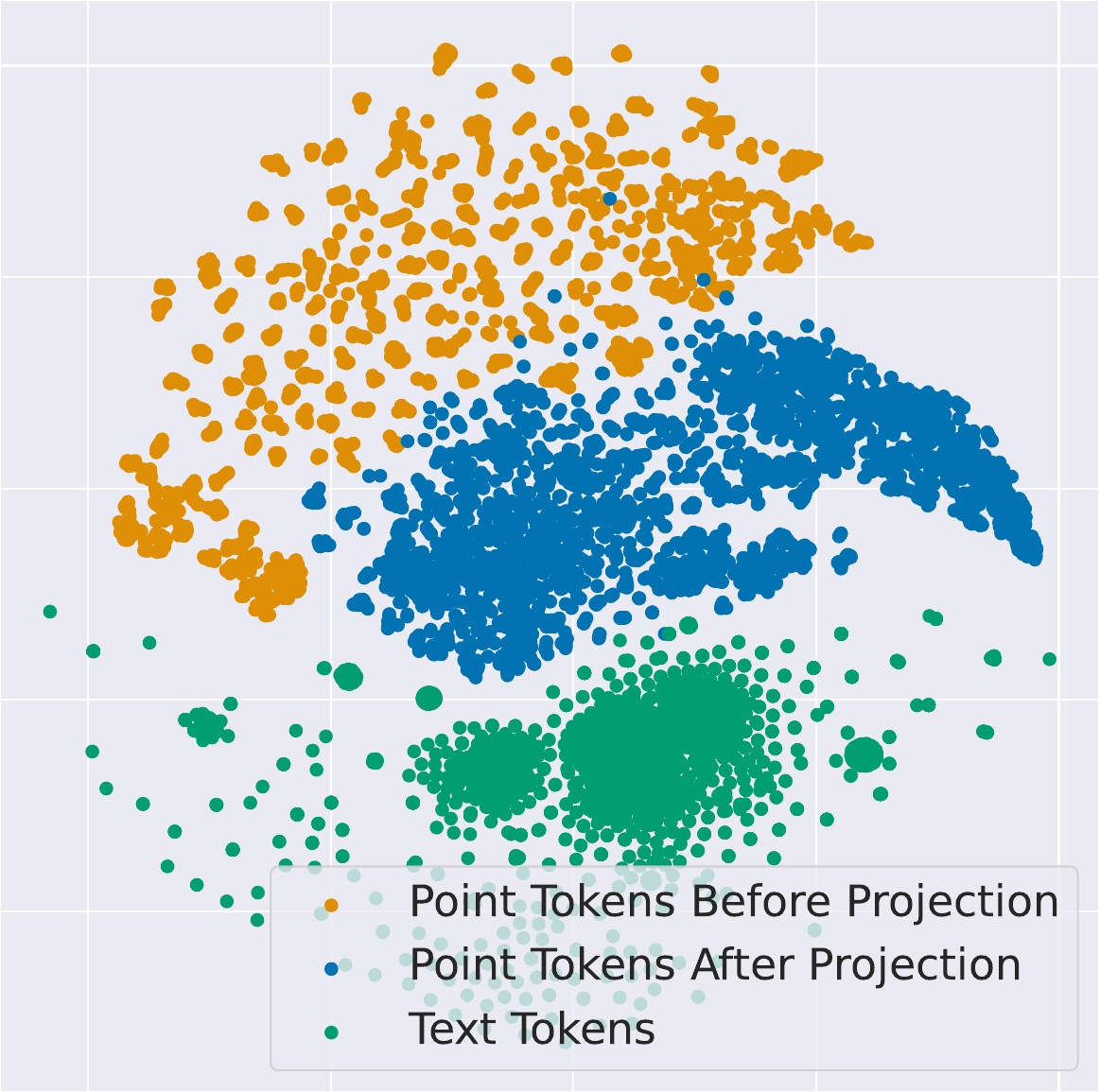}
        \caption{T-SNE visualization of tokens.}
        \label{fig:tsne_visualization}
    \end{minipage}\hfill
    \begin{minipage}{0.53\textwidth}
        \centering
        \includegraphics[width=0.9\linewidth]{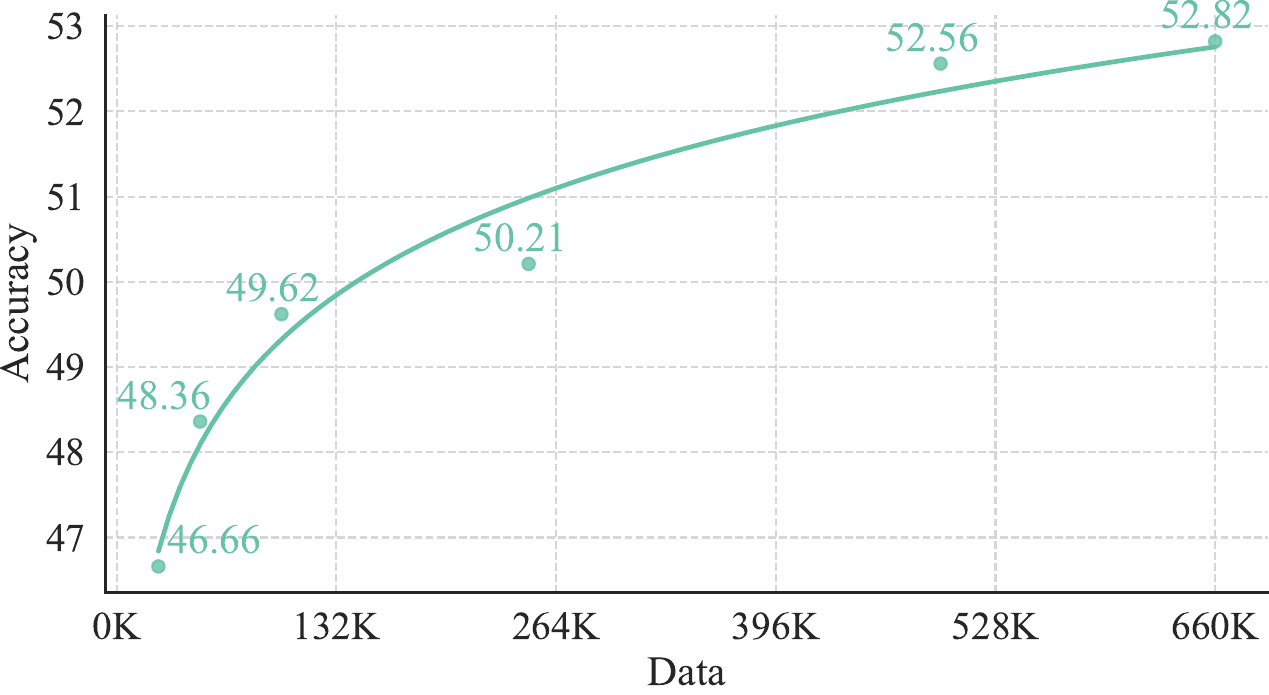}
        \caption{Ablation on data for alignment.}
        \label{fig:alignment_ablation}
    \end{minipage}
\end{figure}
Therefore, we prioritize more comprehensive and reliable measures like human and GPT-4 evaluation along with Sentence-BERT and SimCSE.

\subsection{Ablation Studies}
In this section, we explore various model design choices. Additionally, we examine the impact of different data variations on the training process. The average accuracy of PointLLM on our generative classification benchmark is reported.

\noindent\textbf{Projection layers.}
While the alignment of tokens from different modalities 
to the text space using projection layers is effective and widely used in various domains \cite{LLaVA, MiniGPT-4}, the optimal number of layers required remains an open question. Our experiments, ranging from 1 to 4 projection layers with different hidden dimensions, are detailed in \cref{tab:projection_layer}. Results from both the 7B and 13B models indicate that 3 projection layers yield the best performance. This suggests that both an insufficient and an excessive number of layers can detrimentally affect performance. A balance in the number of layers is thus crucial for optimal model functionality. We also investigate the projection layers' impact by visualizing point tokens' features pre- and post-projection, and text tokens using T-SNE as shown in \cref{fig:tsne_visualization}. Post-projection, point tokens cluster more and align closer with text tokens, verifying the effect of aligning feature spaces. The non-complete overlap may result from the generative, not contrastive, alignment.

\begin{table}[t!]
\centering
{
\hspace{-2em}
\scalebox{0.75}{
\begin{minipage}{0.5\linewidth}
{
\begin{center}
\tablestyle{3.5pt}{1.0}
\caption{\textbf{Projection layers.}}
\label{tab:projection_layer}
\begin{tabular}{@{}ccc@{}}
\toprule
Hidden Dims. & 7B-Acc. & 13B-Acc. \\ \midrule
N.A. & 50.63 & 52.62 \\
1024 & 51.05 & 49.00 \\
\textbf{1024, 2048} & \textbf{52.82} & \textbf{53.39} \\
1024, 2048, 4096 & 52.15 & 51.40 \\ \bottomrule
\end{tabular}
\end{center}
}
\end{minipage}
}
}
\hspace{-3em}
{
\centering
\scalebox{0.75}{
\begin{minipage}{0.5\linewidth}{\begin{center}
\tablestyle{3pt}{1.0}
\caption{\textbf{Max pooling.}}
\label{tab:max_pooling}
\begin{tabular}{@{}ccc@{}}
\toprule
Pooling & Acc. & A100 GPU-Hours \\ \midrule
7B w/ & 48.72 & \textbf{34} \\
7B w/o & \textbf{52.82} & 126 \\
13B w/ & 51.10 & \textbf{56} \\
13B w/o & \textbf{53.39} & 213 \\ \bottomrule
\end{tabular}
\end{center}}\end{minipage}
}
}
\hspace{-2em}
{
\scalebox{0.75}{
\begin{minipage}{0.35\linewidth}{\begin{center}
\tablestyle{3pt}{1.0}
\caption{\textbf{Fine-tuning data.}}
\label{tab:fine-tuning_data}
\begin{tabular}{@{}cccc@{}}
\toprule
Single & Multi. & Detailed & Accuracy       \\ \midrule
\checkmark      &       &          & 40.14          \\
\checkmark      & \checkmark     &          & 45.79          \\
\checkmark      & \checkmark     & \checkmark        & \textbf{52.82} \\ \bottomrule
\end{tabular}
\end{center}}
\end{minipage}
}
}
\label{tab:ablations}
\end{table}

\noindent\textbf{Max pooling.} Unlike sequential or~grid-based text and image tokens, point 
cloud tokens are permutation invariant. Concatenating these tokens with text introduces unnecessary causal dependence and may not be optimal for feature fusion. Inspired by max pooling's symmetric properties \cite{PointNet}, we experimented with aggregating point token information through max pooling before the projection layer. While this method didn't enhance performance as shown in \cref{tab:max_pooling}, it greatly improved efficiency. Training time measured by 80G-A100 GPU-Hour reduced by about 75\%. This underscores the necessity in developing efficient point cloud fusion mechanisms for MLLMs.

\noindent\textbf{Training data.}
To determine the optimal quantity of data for feature alignment, 
we experimented with varying data volumes on our 7B PointLLM, maintaining constant iteration times by duplicating training epochs. Results in \cref{fig:alignment_ablation} suggest that increasing the data volume improves downstream performance, plateauing at around 600K samples. Further, as shown in \cref{tab:fine-tuning_data}, incorporating more types of data during fine-tuning consistently yields performance improvements, underscoring the importance of our diverse instruction-following dataset.

\subsection{Qualitative Results}
\cref{fig:teaser} demonstrates PointLLM's ability to accurately perceive interior details of shoes and cars, overcoming occlusion and viewpoint challenges. More qualitative comparisons of different 13B models are shown in \cref{tab:demo}. Sample 1 from ModelNet40 shows a typical 2D MLLM failure: mistaking a laptop for letters due to depth perception issues inherent in single-view images. While multiple views could potentially alleviate this, they pose challenges in terms of optimal view selection and increased model complexity. Point clouds, however, directly provide object geometry, avoiding issues with depth, occlusion, or viewpoint. Sample 2 highlights PointLLM's capability to generate detailed, accurate captions, outperforming other models and even human annotations, while avoiding severe hallucinations. Notably, despite being trained exclusively on Objaverse, PointLLM can handle scene-level point clouds from the unseen ScanNet\cite{ScanNet} with reasonable outputs, indicating its potential for broader applications. Effectively handling scene-level point clouds necessitates more high-quality data, a resource currently unavailable. We leave it as future work. 
Additional qualitative results in the supplementary material further illustrate the advantages of using point clouds for 3D understanding and PointLLM's superiority.

\begin{table}[t!]
\centering
\caption{\textbf{Qualitative comparisons.} We show the qualitative results of models on ModelNet40\cite{ModelNet40}, Objaverse\cite{Objaverse}, and ScanNet\cite{ScanNet}. Our PointLLM produces more accurate and detailed results than baselines and even human-annotated ground truths.}

\label{tab:demo}
\scalebox{0.88}{\tablestyle{8pt}{1.0}
\begin{tabular}{@{}l p{0.3\linewidth} p{0.4\linewidth}}
\toprule

Samples 1, 2 & 
  \begin{minipage}{\linewidth}
    \includegraphics[width=0.45\linewidth,height=0.4\linewidth]{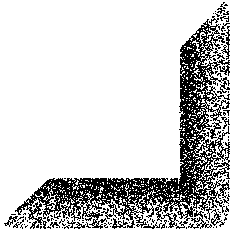}
    \hspace{0.2em}
    \includegraphics[width=0.45\linewidth,height=0.4\linewidth]{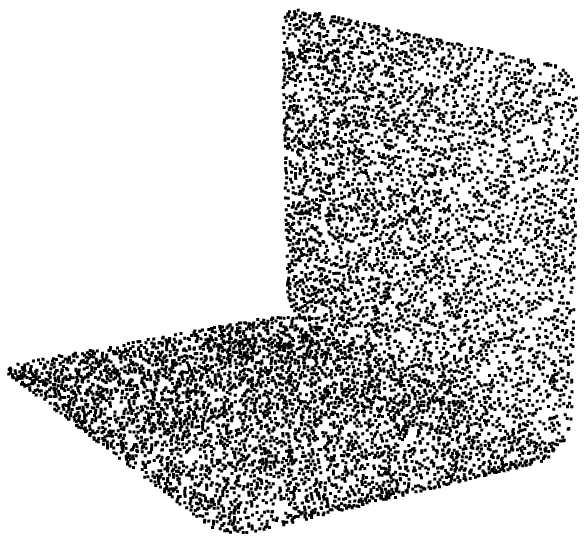}
  \end{minipage}
& 
  \begin{minipage}{\linewidth}
    \includegraphics[width=0.40\linewidth,height=0.3\linewidth]{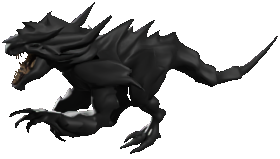}
    \includegraphics[width=0.40\linewidth,height=0.3\linewidth]{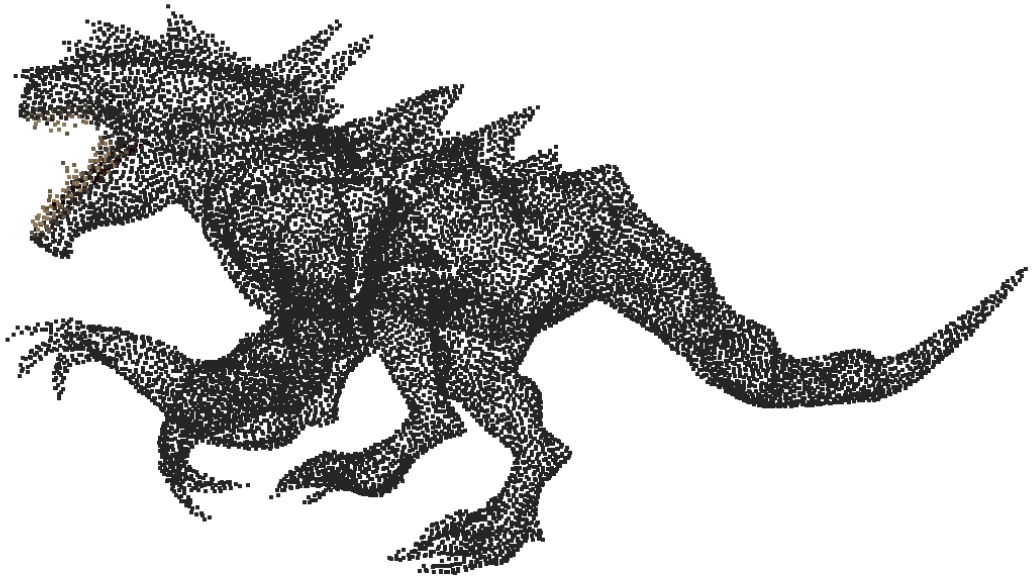}
  \end{minipage}
\\ \midrule
Ground Truth & Laptop & A cartoon black monster like a dragon \\ \midrule

Prompt & What is this? & Briefly caption this 3D model. \\
InstructBLIP\cite{InstructBLIP} & symbol letter l & a black lizard with a sharp tooth in a dark room \\ 
LLaVA\cite{LLaVA} & A small, grainy, black and white letter j. & A 3D model of a dark, menacing dragon. \\ 
3D-LLM\cite{3D-LLM} & - & A black and white tiger with long legs, standing on its hind leg. \\
Point-Bind LLM\cite{Point-Bind} & This is a laptop computer. & The 3D model features a large, ornate gargoyle with a horned helmet, sitting on top of a building. \\
\textbf{PointLLM} & \textbf{The 3D model represents a notebook computer, typically a laptop.} & \textbf{The 3D model depicts a menacing black dragon, with its mouth opened wide revealing a row of sharp teeth.} \\
\bottomrule
\textbf{PointLLM} & \multicolumn{2}{p{0.75\linewidth}}{(The outputs for ScanNet-Scene0611\_00 are shown below.)} \\
\midrule
\begin{minipage}[t]{0.3\linewidth} 
    \raisebox{-\height}{\includegraphics[width=\linewidth]{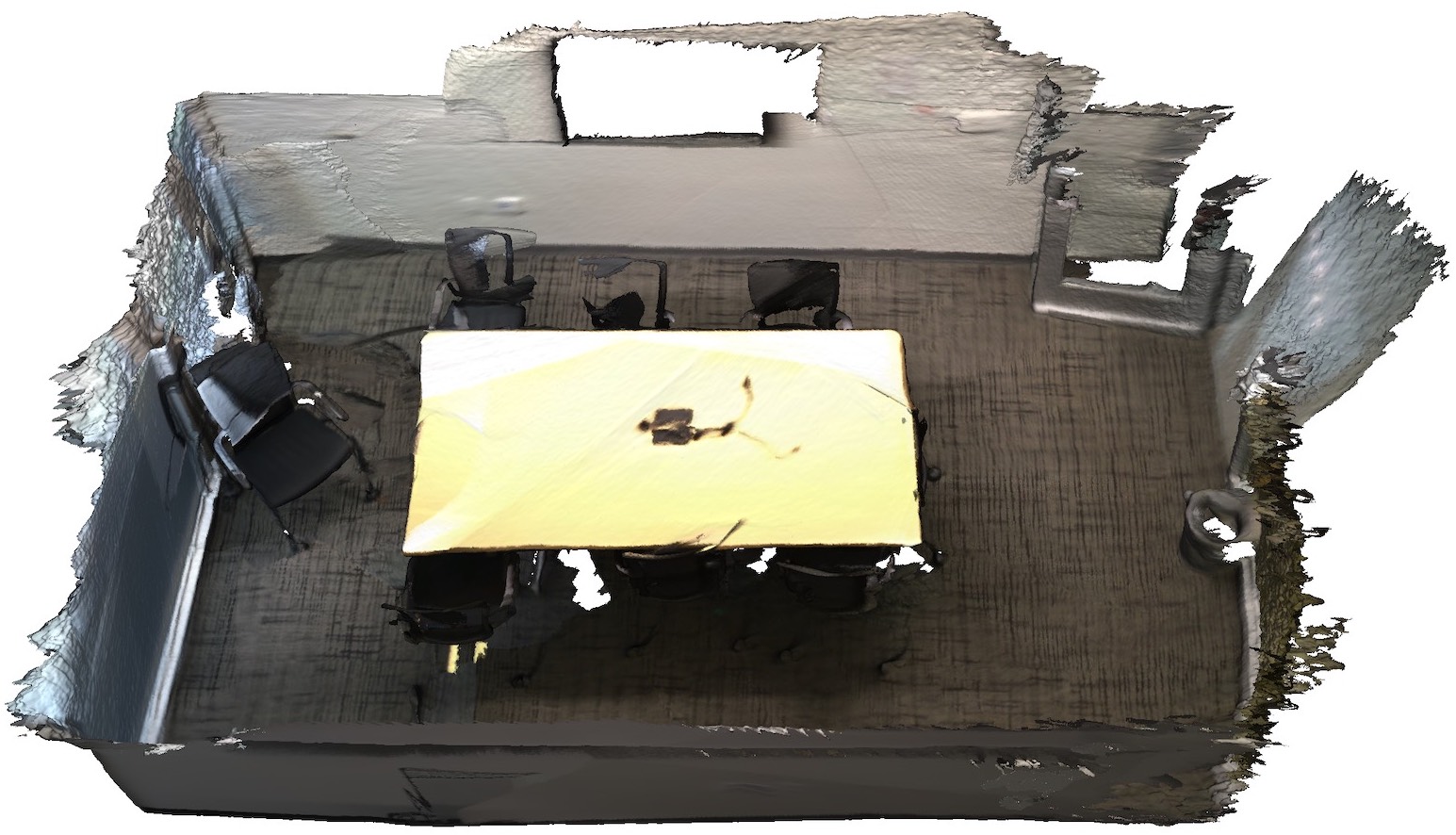}} 
\end{minipage} 
& \multicolumn{2}{p{0.75\linewidth}}{\textbf{This is a 3D model of an interior space in a building, featuring a table and chairs. The room is designed in a minimalistic style, with light-colored walls and dark-colored chairs. The table, presumably made of wood, is the focal point of the room, providing a space for various activities such as dining, study or work. The chairs, looking comfortable and sturdy, suggest a space designed for prolonged sitting.}} \\
\bottomrule
\end{tabular}
}
\end{table}
\section{Conclusions and Future Directions}

In this study, we present PointLLM, a novel and powerful MLLM designed for understanding 3D object point clouds, alongside an automated data generation pipeline and a large-scale dataset. We unveil two innovative benchmarks equipped with a comprehensive evaluation framework, also highlighting the current benchmarks and metrics' limitations. All resources will be open-source. Looking ahead, we aim to refine PointLLM's comprehension of scene-level point clouds and extend its capabilities to include point cloud generation for interactive 3D content creation. Another exciting direction is leveraging PointLLM to generate high-quality 3D object captions at scale, benefiting text-to-3D applications, for which we provide preliminary results in the supplementary material.

\noindent\textbf{Acknowledgements.}\quad We would like to acknowledge Xiangyu Yue for providing feedback about this paper, and thank Lihe Ding, Shaocong Dong, and Jiaming Han for their assistance with the experiments. This research was partially supported by the Centre for Perceptual and Interactive Intelligence (CPII) Ltd. under the Innovation and Technology Commission (ITC)'s InnoHK and Shanghai Artificial Intelligence Laboratory.


%
%
\bibliographystyle{splncs04}
\bibliography{main}



\section*{Supplementary material}

cally generate high-quality 3D object captions for text-to-3D generation. We provide preliminary results related to text-to-3D generation in App. F.

# Appendix

## A. Data Collection

**Instruction lists.** The 30 pre-defined instructions used to prompt the model to briefly and elaborately describe the objects are shown in Tab. 10 and Tab. 11 respectively. These prompts are generated with the assistance of GPT-4 and are coupled with captions to form our description-type data.

**Data generation with GPT-4.** In Tab. 12 we show an example of using GPT-4 for data generation as well as the system prompt of GPT-4. The input is one human-written caption provided by Cap3D[40] and the outputs are one expanded detailed caption, three single-round conversations, and one multi-round conversation. The system prompt is used for all samples, which guides the model to analyze existing captions based on the general knowledge of 3D objects and generate detailed captions, diverse Q&As, and logically connected multi-round conversations.

**Dataset distribution.** The comprehensive statistics of our newly compiled point-text instruction following dataset are detailed in Tab. 9. The dataset encompasses approximately 730K samples. Fig. 5 illustrates the length distributions of instructions and responses across various data types. In Fig. 6, we present word clouds (after removing generic words like "model", "object", etc.) and verb-noun pair distributions following [61] from our instruction following dataset, highlighting its extensive coverage of diverse topics such as color, shape, usage, material, and more.

Table 9. **Statistics of our point-text instruction following data.**

| Statistics | |
|---|---|
| Number of all samples | 731851 |
|   - brief-description type | 661577 |
|   - detailed-description type | 15055 |
|   - single-round type | 40122 |
|   - multi-round type | 15097 |
|     - multi-round responses | 45287 |
| Avg. len. of all instruction/responses (in words) | 11/17 |
|   - brief-description type | 11/15 |
|   - detailed-description type | 9/82 |
|   - single-round type | 10/15 |
|   - multi-round type | 10/21 |

## B. Benchmarks and Evaluation

### B.1. Discussions about Close-Set Classification

Initially, we consider formatting the close-set zero-shot classification task on ModelNet40[64] as a multiple-choice problem, including indexed candidate category names in the prompt, and prompting our model to select one of the 40 categories given the point cloud as input. However, since our model is not designed for multiple-choice problems but for real-world usage where it can generate any word or phrase as output, we cannot directly parse its response for evaluation. Therefore, we use ChatGPT as a post-processor to select one of the ModelNet40 categories based on the model's answer. In the meantime, we find that including category names in the prompt results in meaningless responses from InstructBLIP[9], which is the model we compare with, making meaningful comparisons challenging. Consequently, we opt for a more generalized prompt, without including the candidate lists in the prompt. This allows us to make balanced comparisons.

Including candidate lists in the prompt, we also tried to calculate the conditional probability of different options given the model's output following [62], but this method did not work well for our model. As our instruction-following training data lacks such scenarios where it's needed to choose from a fixed set of options, our model always produces very low probabilities on these options with biased results. For example, among the options "00" to "39", our model predicts very low probabilities and among these low probabilities, "00" and "39" are the highest most of the time, which leads to biased predictions. Therefore, we choose to use general prompts and utilize ChatGPT for post-processing. This approach more accurately reflects real-world scenarios where the model is expected to provide natural, free-form responses to diverse, unstructured questions.

## B.2. GPT Evaluation Prompts

**Close-set zero-shot classification.** In this task, we use ChatGPT to post-process the model output by selecting the most probable class index from the 40 ModelNet40 categories. The process is detailed in Tab. 13, where {candidate_lists} refers to the ModelNet40 category list, and {model_output} refers to the model's response. ChatGPT is required to directly output the category index, category name, and a short reason for the choice. If the description doesn't clearly refer to any one of the categories, ChatGPT must make an educated guess based on the information provided. If ChatGPT cannot infer, then "-1" is returned and a random index will be chosen as the model's classification prediction. We do not use a system prompt for ChatGPT but directly input the prompt.

**Open-vocabulary classification.** In this task, we use GPT-4 as an evaluator to classify whether the model's response and the human caption are referring to the same object type. The process is outlined in Tab. 14, where {ground_truth} and {model_output} refer to the human caption and the model's response. We do not require the model's response to match exactly with the human caption, as long as it conveys the same object type. We also directly input the prompt for GPT-4 instead of using a system prompt.

**Object captioning.** In this task, we utilize GPT-4 as an evaluator to assess model-generated captions against human-generated captions (ground truth) of 3D models. GPT-4 is tasked with identifying aspects mentioned in the human caption and calculating the percentage of these aspects that are either correctly mentioned or partially matched in the model's caption on a scale of 0 to 100, with each aspect contributing equally to the score. The evaluation process is detailed in Tab. 15, where {ground_truth} refers to the human caption, and {model_output} refers to the model's response.

## B.3. Human Verification of GPT Evaluation

To verify the effectiveness of using GPT models for evaluation, the first author manually checks the evaluation results of ChatGPT and GPT-4.

In the close-set classification task on ModelNet40, the author finds the following:

1. ChatGPT consistently outputs in the desired format, selecting the category or "-1" and providing a reason.
2. When the model output clearly refers to or hints at a category with salient information regarding one of the candidate categories, ChatGPT can accurately identify the corresponding category based on the model's output, showing a high degree of consistency with human-selected options. False negatives or false positives are rare in these cases.
3. If the model output is ambiguous, ChatGPT's selection appears random, aligning with our expectations for handling such cases in classification tasks, because when the model encounters uncertainty or lacks confidence in its identification, random guessing is permissible.

For open-vocabulary classification and object captioning tasks on Objaverse, the author finds that ChatGPT underperforms in identifying the same object concept, acting as a strict judge, and producing more false negatives in classification. It often considers two words or phrases not to refer to the same object type, even when they do. In contrast, GPT-4 demonstrates accurate recognition. After reviewing 50 samples of classification results, the first author has 100% consistency with GPT-4's evaluations. As a result, we opt to use GPT-4 for the open-vocabulary and object captioning tasks on Objaverse. Examples of GPT evaluation can be found in Tab. 13, Tab. 14, and Tab. 15.

## B.4. Human Scoring Criteria

Human evaluators were employed to assess captions in the object captioning benchmark. Outputs for the same object from various models were grouped and randomly shuffled, and evaluators independently scored these captions while manually inspecting objects in the Objaverse using the official explorer at https://objaverse.allenai.org/explore.

**Scoring criteria.** The evaluation process involved assigning correctness scores and hallucination scores following these guidelines:

1. **Correctness score.**
   - Each distinct correct attribute in a model output (*e.g.*, category, color, shape, usage, material) was awarded one point. For example, a black tire correctly identified as a tire and being black would receive two points.
   - Partial correctness was graded on a scale of 0 to 1, depending on the degree of accuracy. For instance, if a model output described "a cartoon figure" but the object was specifically a cartoon horse, it would be awarded 0.5 points.
2. **Hallucination score.**
   - Hallucination points were assigned for each incorrect detail in the model output, mirroring the correctness scoring mechanism. For instance, if the model incorrectly described two yellow tires instead of four black ones, it would incur two hallucination points, one for color and one for number.
   - Repetitive inaccuracies based on one attribute were not subject to multiple penalties. As an example, erroneously mentioning a black tire when no tire existed would lead to only one hallucination point.
   - Penalties were also applied for content that was irrelevant to the object description.
3. **General considerations.**
   - Generic terms like '3D model' or 'image' were disregarded, as were references to black backgrounds or environmental colors, and viewpoints.

- Elements that were indeterminable as either correct or incorrect were not considered in the scoring.
- Within each group of evaluations, a range of scores should be established to differentiate between high and low-quality captions. After the initial scoring, a final review adjusted the scores to ensure a clear distinction between better and worse captions.

**Precision score.** We also calculate the precision score as the proportion of correct information within the model-generated content as follows:

$$\text{Precision} = \frac{\text{C. Score}}{\text{C. Score} + \text{H. Score}} \times 100\% \qquad (3)$$

where **C. Score** refers to the correctness score and **H. Score** refers to the hallucination score. The precision score in the main paper is reported after summing all **C. Score** and **H. Score** of all samples for robust evaluation.

## C. Implementation and Training

**Implementation details.** We use ULIP-2[66] to pre-train our point cloud encoder (Point-BERT[68]). ULIP-2 is a method for aligning the latent space of the point cloud encoder to that of CLIP[49] through contrastive learning, endowing the encoder with a strong zero-shot capability for 3D object recognition. As the original implementation of ULIP-2 only supports point clouds with spatial coordinates (xyz), we re-train Point-BERT from scratch with color information (xyzrgb), following the same procedure outlined in the ULIP-2 paper. For training Point-BERT, we employ ViT-L/14 trained on DataComp-1B[13] for 12.8B steps and batch size 90k (denoted as 'ViT-L/14-datacomp_xl_s13b_b90k') from OpenCLIP[29]. We use point clouds from Cap3D[40], which contains 660K objects. We filter out 3000 objects from this dataset and reserve them for future testing. These 3000 objects are not used during any stage of the entire model training and the 200 objects utilized for our benchmarks are part of these 3000 unseen objects to prevent information leakage.

**Training details.** All training are conducted on 8 × 80G A100 GPUs with BF16 data type, leveraging flash-attention [10], the AdamW [39] optimizer, and a cosine learning rate scheduler. For the feature alignment stage, we train our model for 3 epochs with a batch size of 128 and a learning rate of 2e-3. For the instruction tuning stage, we train our model for 3 epochs with a batch size of 32 and a learning rate of 2e-5. For efficiency, the 7B model completes the feature alignment and instruction tuning stages in approximately 13.3 and 2.5 hours, respectively, while the 13B model takes around 22.3 and 4.3 hours for the same stages.

## D. Win Rate Comparison

In Fig. 7, we present the win rate comparisons of the 7B and 13B model variants across different scores. Note that due to

truncation errors in plotting figures, some cumulative rates (win, tie, lose) may not sum to exactly 100%.

The win rate analysis reveals that both the 7B and 13B models not only outperform baselines but also surpass human annotators in terms of correctness scores by a significant margin. This superior performance in correctly identifying object attributes underscores the models' advanced understanding and processing capabilities of 3D objects.

Regarding hallucination, our models exhibit a marked improvement over the baselines, reflecting their better ability to avoid generating incorrect or fabricated details about the objects. This is further corroborated by the precision scores, where our models demonstrate a higher ratio of correct information in their outputs compared to other models. Compared with human annotators, our models show comparable performance in about 50% of samples in terms of hallucination and precision.

However, it must be acknowledged that there is still room for improvement in reducing hallucination rates to match the levels achieved by human annotators. Striving towards the precision demonstrated by human evaluations remains a target for future enhancements.

## E. Qualitative Results

In this section, we provide the qualitative results from different datasets of the 13B models for comparison. All samples used were unseen by our models during training.

**Results on ModelNet40.** Tab. 16 illustrates the classification results from different models on the ModelNet40 dataset. These examples highlight the inherent limitations of image-based models, which depend on suitable views for accurate object identification. Notable challenges include the failure to recognize the guitar in Sample 3, the monitor in Sample 6, and the ambiguity in depth perception leading to misclassifications, such as confusing a chair with a bed in Sample 2, and a bathtub with a bowl in Sample 5. In contrast, PointLLM bypasses these challenges by using point clouds, which provide direct access to object geometry without concerns over ambiguous depth, occlusion, or viewpoint.

Moreover, there are evident hallucination issues with other models. For instance, Point-Bind LLM[18] erroneously describes a person lying on the couch in Sample 1 and someone holding a wine bottle in Sample 4. In comparison, our 13B model consistently provides accurate and realistic classifications. Its superior performance, devoid of such hallucinatory inaccuracies, underscores the model's advanced comprehension of 3D structures and its effectiveness in handling diverse object types.

**Results on Objaverse.** Due to the limited capability of Point-Bind LLM in producing meaningful outcomes, it has been excluded from our comparative analysis on Objaverse. It is noteworthy that InstructBLIP also occasionally yields

nonsensical results as in Sample 2 of Tab. 17. As depicted by the results, PointLLM consistently generates captions that are both more accurate and detailed compared to other baselines and human annotators. For example, in Sample 2, PointLLM accurately describes the golden brown eyes of an insect, a detail overlooked by human annotators who provide only a generic description, and completely missed by other models failing to identify the object type correctly. Similarly, in Samples 3 and 4, PointLLM offers elaborate descriptions encompassing shape and color without errors, in stark contrast to the simplistic captions from human annotators and erroneous information from other baselines.

**Dialogues.** Fig. 8 showcases dialogues between PointLLM and a human user, which reveal PointLLM's capacity to understand point clouds' shapes, appearances, functionalities, and more. Notably, our PointLLM is unaffected by occlusion, capable of discerning the car's internal two-seat structure and identifying a logo on the back of a shoe, tasks challenging for image inputs. Furthermore, our model engages with human instructions using common sense and avoids biases, as seen in its refusal to declare a 'best' shoe brand. Collectively, these samples validate PointLLM's proficiency in understanding point clouds and responding to human instructions both accurately and effectively.

## F. Text-to-3D Generation

The burgeoning interest in text-to-X generation tasks [5, 31, 41, 52] has led to significant advancements. Notably, [5] demonstrated that text-to-image generation models benefit greatly from training on highly descriptive, generated captions. Leveraging PointLLM's capability to generate detailed and accurate captions for 3D models, we explore its potential in enhancing text-to-3D generation models.

We employed PointLLM-13B to generate captions for Objaverse[11] objects with LVIS[19] labels. The prompt "Describe this 3D model in detail and accurately." was used for caption generation. We trained the text-to-3D generation model from [2] using captions generated by our PointLLM and the Cap3D[40] respectively for comparison.

Qualitative comparisons of the generation results, as illustrated in Fig. 9, reveal that the model trained with PointLLM-generated captions generates objects more closely aligned with text prompts and exhibits more precise detailing. This highlights the advantages of utilizing detailed and accurate captions from our model for text-to-3D generation tasks, pointing towards improved fidelity and coherence in generated 3D objects.

(a) **Brief description-instruction.**

(b) **Brief description-response.**

(c) **Detailed description-instruction.**

(d) **Detailed description-response.**

(e) **Single round-instruction.**

(f) **Single round-response.**

(g) **Multi round-instruction.**

(h) **Multi round-response.**

Figure 5. **Length distributions of instructions and responses of different types of point-text instruction following data.**

(a) **Brief description-instruction.**



(b) **Brief description-response.**



(c) **Detailed description-instruction.**



(d) **Detailed description-response.**



(e) **Single round-instruction.**



(f) **Single round-response.**



(g) **Multi round-instruction.**



(h) **Multi round-response.**

Figure 6. **Word distributions of instructions and responses of different types of point-text instruction following data.**

14

Table 10. **The instruction list for brief descriptions.** An instruction from the list is randomly selected and coupled with a human-written caption from Cap3D[40] to form a brief-description instruction following sample.

- Summarize the 3D point cloud object briefly.
- What kind of object is depicted by this point cloud?
- Provide a short explanation of this 3D structure.
- What does this collection of points represent?
- Offer a succinct summary of this 3D object.
- Can you give a brief overview of this point cloud?
- Characterize the object this point cloud is illustrating.
- Share a brief interpretation of this 3D point cloud.
- Provide an outline of this 3D shape's characteristics.
- What object is this point cloud rendering?
- Deliver a quick description of the object represented here.
- How would you describe the 3D form shown in this point cloud?
- What is the nature of the object this point cloud is representing?
- Present a compact account of this 3D object's key features.
- What can you infer about the object from this point cloud?
- Offer a clear and concise description of this point cloud object.
- How would you summarize this 3D data set?
- Give a brief explanation of the object that this cloud of points forms.
- What kind of structure does this 3D point cloud depict?
- Could you delineate the form indicated by this point cloud?
- Express in brief, what this point cloud is representing.
- Give a quick overview of the object represented by this 3D cloud.
- Convey a summary of the 3D structure represented in this point cloud.
- What kind of object is illustrated by this collection of points?
- Describe the object that this point cloud forms.
- How would you interpret this 3D point cloud?
- Can you briefly outline the shape represented by these points?
- Give a concise interpretation of the 3D data presented here.
- Explain the object this point cloud depicts succinctly.
- Offer a summary of the 3D object illustrated by this cloud.

Table 11. **The instruction list for detailed descriptions.** An instruction from the list is randomly selected and coupled with a GPT-4 generated caption to form a detailed-description instruction following sample.

- Can you tell me more about this?
- What does this represent?
- Can you describe this in more detail?
- I'm interested in this, can you explain?
- What is this object made of?
- Could you provide more info about this?
- What exactly am I looking at here?
- What is this?
- Could you describe the detailed structure of this?
- This looks interesting, can you expand on it?
- Can you explain more about this form?
- What can you tell me about the shape of this object?
- Could you delve deeper into this?
- I want to know more about this, can you help?
- Can you walk me through the details of this object?
- Can you provide a comprehensive account of this object?
- Offer a detailed interpretation of this point cloud.
- Please elucidate on the characteristics of this form.
- Could you provide an in-depth description of this structure?
- What does this cloud represent in its entirety?
- Elaborate on the details of this point cloud, please.
- Kindly furnish me with more information about this object.
- Please expand on the intricate structure of this form.
- Provide a meticulous explanation of what these points represent.
- I request a detailed breakdown of this structure.
- Give a thorough rundown of this point cloud.
- Can you offer a complete analysis of this object?
- I would like a comprehensive explanation of this form.
- Please detail the specific features of this point cloud.
- Could you elaborate extensively on what this represents?

Table 12. **An example of data generation with GPT-4.** The input consists of a human-written caption provided by Cap3D[40], guided by a system prompt to analyze the existing caption based on the general knowledge of 3D objects. The outputs include an expanded detailed caption, three single-round conversations, and one multi-round conversation.



| | |
|---|---|
| Uid | 4887da0aab51406dab3c5cb69ec82404 |
| System Prompt | Analyze the 3D object model from the given caption:<br>1. Write a new detailed caption by describing it in 50-100 words, identifying its type, appearance, and any other inferred information such as functionalities, usages, daily-life knowledge, etc. Exclude any uncertain details.<br>2. Generate 3 single-round Q&As regarding diverse aspects of the object based on the provided captions and your new captions.<br>3. Construct 1 set of 3-round Q&As, ensuring logical relevance between the questions and answers. Questions should be different from those in `"2"`.<br><br>Format your response as:<br>` ` ` json<br>{<br>    "caption": "description",<br>    "single_conversation": [{"Q": "Q", "A": "A"} x 3],<br>    "multi_conversation": [{"Q1": "Q", "A1": "A", "Q2": "Q", "A2": "A", "Q3": "Q", "A3": "A"} x 1]<br>} |
| Input | A female doll with long back cream hair and with black high boots. |
| Output | {<br>    "caption": "The 3D object model represents a detailed doll figure of a female character with long, cream-colored hair cascading down her back. It is dressed in fashionable attire with black high boots being one of its highlighted features. The doll, resembling those used in play or as collectible items, exhibits a high degree of intricate detailing, showcasing its fashionable outfit and engaging features, making it appealing for users, especially ones intrigued by doll collection or imaginative play.",<br>    "single_conversation": [<br>        {"Q": "What color is the doll's hair?", "A": "The doll's hair is cream-colored."},<br>        { "Q": "What is specific about the doll's footwear?", "A": "The specific footwear feature of the doll is its black high boots."},<br>        {"Q": "What are possible usages of this doll?", "A": "Possible usages for this doll might include being part of a toy collection, used for imaginative play, or as a detailed model for hobbyists or artists."},<br>    ]<br>    "multi_conversation": [<br>        {"Q1": "What makes this doll distinguishable?", "A1": "The doll is distinguishable by its long cream-colored hair cascading down its back and its outfit featuring black high boots.", "Q2": "Could this doll be part of a collection or series?", "A2": "Yes, based on the intricate detailing and appearance, it's possible that the doll could be part of a collection or series.", "Q3": "Who would be the likely audience or users of this doll?", "A3": "The likely users of this doll would be individuals who enjoy doll collecting, children or adults for imaginative play, or even hobbyists and artists who may use it as a model."}<br>    ]<br>} |

Table 13. **The prompt and examples of ChatGPT in close-set zero-shot classification.** ChatGPT post-processes the model output and selects the most probable class index from the available 40 categories, even if the description is vague or ambiguous. The blue placeholders {candidate_lists} and {model_output} refer to the ModelNet40 category list and the model's response, respectively.

| Prompt | Given the following free-form description of a 3D object, please determine the most probable class index from the following 40 available categories, even if the description doesn't clearly refer to any one of them. Make your best-educated guess based on the information provided. If the description already contains a valid index, then the index should be selected. If it contains more than one valid index, then randomly select one index (specify your reason). If there is no valid index and it cannot be inferred from the information, return "-1#NA#Cannot infer". <br> Categories: <br> 0: airplane <br> 1: bathtub <br> {candidate_lists} <br> 38: wardrobe <br> 39: xbox <br> Reply with the format of "index#class#short reason (no more than 10 words)". <br><br> Examples: <br> Input: This is a 3D object model of a cartoon white truck. <br> Output: 7#car#Closest match to "car" in categories. <br><br> Input: A green leaf in a flower pot. <br> Output: 26#plant#The primary subject "leaf" directly indicates a plant. <br><br> Input: It's difficult to determine the exact type of this object due to insufficient details. But it seems to be like a piece of furniture. <br> Output: 33#table#Randomly select one kind of furniture from the list. <br><br> Input: I cannot determine the specific type of the object without additional information or context. <br> Output: -1#NA#Cannot infer. <br><br> Now analyze the following: <br> Input: {model_output} <br> Output: |
|---|---|
| Example 1 | Input: This 3D model represents a war plane, typically used for combat and military purposes. It is characterized by a large, dominating hue of military gray. The detailing on the model suggests it's a fighter jet, hinting at its usage in air combats. Its posture seems ready for flight, showcasing a typical attribute of war planes - preparedness for battle. This could be used in game design or other virtual reality applications that require detailed, lifelike renderings of objects. <br> Output: 0#airplane#The description explicitly mentions "war plane" and "fighter jet". |
| Example 2 | Input: The 3D model object is a tall, cylindrical-shaped clay container with a flat, smooth top and a slightly wider base. The cylindrical form and the structure suggest that it's possibly used for storing liquids or semi-liquids due to its wide top and flat base allowing easy pouring and stability. Its material and structure indicate it could be a common household item, like a vase or a jug. <br> Output: 37#vase#The description includes key characteristics of a vase (tall, cylindrical shape, wide top, flat base). |
| Example 3 | Input: This is a 3D model of a sleek, black rectangular unit with its sides elegantly curved. On one of its sides, there is a distinctive circle inscribed, adding a unique character to the model. The object's form and design suggest that it can be a piece of technology, like a gadget or a home appliance, with the circle possibly indicating a button or a sensor. Its usage, however, cannot be ascertained without specific context. <br> Output: 22#monitor#The description suggests a sleek, black rectangular unit which can be a monitor. The inscribed circle may indicate a power button or a sensor on the monitor. |

Table 14. **The prompt and examples of GPT-4 in open-vocabulary classification.** GPT-4 needs to analyze two sentences to determine if they refer to the same general object or concept, focusing on the type of object, not attributes such as color, size, or shape. The placeholders {ground_truth} and {model_output} refer to the human caption and the model's response, respectively.

| | |
|---|---|
| Prompt | Analyze two sentences and determine if they're referring to the same general object or concept, focusing on the type of object, not attributes such as color, size, or shape. Respond with `"T"` if they refer to the same thing and `"F"` if not. Also, provide a brief rationale (no more than 20 words) for your judgment.<br>Example:<br>Input: 1. Spiral staircase that goes from a ground floor. 2. This is a 3D model of wooden stairs in light brown<br>Output: T#Both refer to a staircase.<br><br>Now, analyze the following:<br>Input: 1. {ground_truth} 2. {model_output}<br>Output: |
| Example 1 | Input: 1. A black and brown colored gun. 2. The 3D object is a representation of a futuristic, high-tech gun crafted from a glossy black material. Distinctive features include its metallic handrail, giving an impression of a robust mechanized design. The gun, possibly used in a sci-fi or futuristic setting, denotes advanced technology and might include functionalities such as voice recognition, aiming systems, or biometric triggers.<br>Output: T#Both refer to a gun. |
| Example 2 | Input: 1. A yellow and white fish with black stripes and fins. 2. This is a 3D model of a vibrant, polka-dotted toy fish that is predominantly orange on the body, shifting to white on the belly. The toy has dark brown spots that enhance its appearance, potentially mimicking the natural patterns found on real-life fish. It's an ideal object for educational purposes, helping to introduce children to marine life, as well as serving as a playful item in a playroom or nursery.<br>Output: T#Both refer to a fish. |
| Example 3 | Input: 1. A white cartoon scorpion with eight legs. 2. This is a 3D object model representing a cartoon version of a rare type of spider. The entire model is rendered in white, which highlights its unique and exaggerated characteristics such as multiple legs and a funnel-like body. Its cartoonish appeal makes it more appealing to a younger audience, and it could possibly be used in animations or educational materials to teach children about spiders in a less intimidating way.<br>Output: F#One is a scorpion and the other is a spider. |

Table 15. **The prompt and examples of GPT-4 in object captioning.** GPT-4 evaluates the model's response by identifying aspects mentioned in the human caption and calculating the percentage of aspects that are correctly or partially matched in the model's caption. The placeholders {ground_truth} and {model_output} refer to the human caption and the model's response, respectively.

| | |
|---|---|
| Prompt | Evaluate a model-generated caption against a human-generated caption (ground truth) for a 3D model. Identify the aspects mentioned in the human caption and calculate the percentage of these aspects correctly mentioned or partially matched in the model caption. Score from 0 to 100, where each aspect contributes equally to the score. Consider similar concepts for a partial score.<br><br>Provide your score (0-100) and a short justification (less than 15 words) in the format of "score#reason"<br><br>Example:<br>Human: A white brown skeleton<br>Model: This is a 3D model of a small, cartoon-like robot. It has a spherical body and is covered in a layer of white dust.<br>Output: 50#mention white; skeleton and robot have similar appearence.<br><br>Now score the following:<br>Human: {ground_truth}<br>Model: {model_output}<br>Output: |
| Example 1 | Human: A white presentation of a planted trees plantation region.<br>Model: The model is a representation of a table saw with a simple, four-legged design.<br>Output: 0#No aspects from the human caption are mentioned in the model's. |
| Example 2 | Human: Private jet<br>Model: This is a 3D model of a cartoon-style airplane.<br>Output: 50#mentioned airplane, but missed private jet aspect. |
| Example 3 | Human: A cartoon look like a tree in 3d<br>Model: The 3D model is an animated depiction of a tree with gray bark, characterized by its lifelike details and textured surface that mimic the realistic grain of bark.<br>Output: 100#mentioned tree, 3D and artistic/cartoonish aspect. |

Figure 7. **Win rate comparisons of 7B and 13B models across various scores.** Our models surpass baselines and human annotators in correctness scores, while exhibiting lower hallucination rates and superior precision compared to various baselines.

Table 16. **Qualitative results on ModelNet40.** The first image in each sample serves as the input for image-based models, with additional point cloud views provided for reference. PointLLM consistently and accurately identifies object types, whereas other models struggle with correct identification or produce hallucinated, incorrect content.

| Samples 1, 2 |  | |  |
|---|---|---|---|
| Ground Truth | Sofa | | Bed |
| Prompt | What is this? | | This is an object of |
| InstructBLIP[9] | sofa | | person cutting meat into small pieces with a large knife |
| LLaVA[37] | The image is a black and white drawing of a couch. | | The image is a black and white depiction of a chair. |
| Point-Bind LLM[18] | This is a drawing of a person laying on a couch, with a dog nearby. | | This is an object of art. |
| **PointLLM** | **This 3D object is a comfortable, black, three-seater sofa.** | | **This is a model of a neatly made bed, primarily composed of a solid black mattress.** |

| Samples 3, 4 |  | |  |
|---|---|---|---|
| Ground Truth | Guitar | | Bottle |
| Prompt | What is this? | | This is an object of |
| InstructBLIP[9] | telescope | | wine bottle |
| LLaVA[37] | The image is a close-up view of a black ink pen sitting on a white surface. | | This image features a black and white depiction of a bottle. |
| Point-Bind LLM[18] | This is a bird flying in the sky. | | This is an object of a person holding a bottle of wine. |
| **PointLLM** | **This 3D model represents a black electric guitar equipped with a distinctive headstock.** | | **This is a 3D model of a bottle, possibly intended for liquids.** |

| Samples 5, 6 |  | |  |
|---|---|---|---|
| Ground Truth | Bathtub | | Monitor |
| Prompt | What is this? | | This is an object of |
| InstructBLIP[9] | table | | the foot |
| LLaVA[37] | The image features a black and white, close-up view of a grater, which is an object used for grating cheese or other food items. | | The image features a small table with a black base and a white or light-colored top. |
| Point-Bind LLM[18] | This is a bowl or a vase, which is oval-shaped and has a smooth surface. | | This is an object of art, a sculpture, or a decorative piece. |
| **PointLLM** | **The 3D object model represents a realistic, dark grey bathtub.** | | **This is a 3D model of a desktop computer monitor.** |

Table 17. **Qualitative results on Objaverse.** The first image in each sample serves as the input for image-based models, with additional point cloud views provided for reference. PointLLM can generates more accurate and detailed captions for 3D objects than other baseline models and human annotators.

| Samples 1, 2 |  |  |
|---|---|---|
| Uid | 0ea33b6617174530b97d6b7a92c275fb | de8ec2a724f14fc4b54624512f80f13e |
| Prompt | What is this? | This is an object of |
| Human | A cartoon green and red like a fruit | A black insect |
| InstructBLIP[9] | an appleavatar 3d model | 26 |
| LLaVA[37] | This is a computer-generated image (CGI) or a 3D model of an apple with a leaf on it. | This is an image of a large, close-up, and dark-colored insect, possibly a big cockroach, with long antennae. |
| 3D-LLM[25] | A 3D model of a red apple. | A small, black spider with a long tail. |
| **PointLLM** | **This is a 3D model of a unique apple, distinctively adorned with a single, vibrant green leaf at the top.** | **This 3D model depicts a realistic, jet-black insect with a pair of striking, golden brown eyes.** |

| Samples 3, 4 |  |  |
|---|---|---|
| Uid | 059a7936ed89419ba9eae3153753ae86 | 82413f74efb84bc3b872a7a8f26cca81 |
| Prompt | Caption this 3D model in detail. | Caption this 3D model in detail. |
| Human | A blue cartoon | A white fish with orange/red spots |
| InstructBLIP[9] | There is a cute blue and white pony in a still, looking cheerful and cheerful. It's standing next to a larger blue-colored object, potentially a barrier or structure, which provides the backdrop for this scene. The overall image is designed to create a fun, whimsical setting for the pony. | There is a fish hanging on a string in front of a black background. The fish is swimming in the middle of the image and is a little bit up from the bottom of the screen. The fish has a red spots on the left and right sides, and is positioned slightly towards the top of the screen. |
| LLaVA[37] | The image features a 3D model of a cute, colorful cartoon unicorn with a blue and white body and a hint of pink. The unicorn has a magical aura and is set against a blue background. Its long, flowing mane and tail add to its whimsical charm. The unicorn appears to be in motion, possibly flying or prancing, as it captures the viewer's imagination. | The image features a 3D model of a fish floating in a dark, empty space. The fish is positioned towards the left side of the image and appears to be swimming upwards. The fish is large and occupies a significant portion of the image, giving the impression of a life-like, computer-generated representation. The dark background and lack of other objects in the scene emphasize the fish's presence and create a sense of depth and dimension. |
| 3D-LLM[25] | A 3D model of a purple and pink pony with long hair, standing on its hind legs. | A small fish with a long tail is visible in the image. It has no distinguishing features or markings, and its color appears to be brown. |
| **PointLLM** | **A 3D model of a vibrant blue cartoon pony, characterized by its large white eyes, two wings, and a distinctive horn. The creature's overall design exudes a playful and friendly aura, making it appropriate for children's animations or toy production. The noticeably large white eyes and the unique wing and horn features endow the pony with an element of fantasy, while its blue body color gives it an overall appealing look.** | **The 3D model is of a long fish, illustrating a striking color combination of red, yellow, and grey. It's characterized by a slender tail. The yellow color is mostly seen on the body of the fish, fading into a subtler, grayish hue towards the fins and tail. The model depicts a typical long fish species, often found in the ocean, and can be used in a variety of digital applications, like animation, games, or virtual reality.** |

**Figure 8. Dialogues between PointLLM and a human user.** The dialogues show PointLLM's ability to understand point clouds' shapes, appearances, functionalities, etc. PointLLM's resilience to occlusion enables it to perceive detailed features such as the interior structure of a car or a logo on a shoe, which are challenging for image inputs. Additionally, PointLLM demonstrates abilities to respond to human instructions with common sense, avoiding biases.

24

(a) **PointLLM.**

(b) **Cap3D.**

Text prompt: "a chimpanzee stirring a bubbling purple potion in a cauldron."

(c) **PointLLM.**

(d) **Cap3D.**

Text prompt: "a chihuahua wearing a tutu."

(e) **PointLLM.**

(f) **Cap3D.**

Text prompt: "a confused beagle sitting at a desk working on homework."

Figure 9. **Text-to-3D generation results of models trained with different captions.** The Model trained with PointLLM-generated captions generates objects more closely aligned with text prompts and exhibits more precise detailing.



\end{document}